\documentclass[10pt,twocolumn,letterpaper]{article}
\pdfoutput=1
\usepackage{iccv}
\usepackage{times}
\usepackage{epsfig}
\usepackage{graphicx}
\usepackage{amsmath}
\usepackage{amssymb}
\usepackage{mathtools}
\usepackage{multirow}
\usepackage{booktabs}
\usepackage{hhline}
\usepackage{ragged2e}
\usepackage{tabularx}

\iccvfinalcopy 

\def\iccvPaperID{5829} 

\DeclarePairedDelimiter\floor{\lfloor}{\rfloor}

\newcounter{savecntr}
\newcounter{restorecntr}
\begin{document}

\title{SF-Net: Structured Feature Network for Continuous Sign Language Recognition}

\author{
\begin{tabular}{c}
Zhaoyang Yang$^{1}$\setcounter{savecntr}{\value{footnote}}\thanks{Equal contribution},\space\space Zhenmei Shi$^{2}$\setcounter{restorecntr}{\value{footnote}}\setcounter{footnote}{\value{savecntr}}\footnotemark \setcounter{footnote}{\value{restorecntr}},\space\space  Xiaoyong Shen$^1$,\space\space  Yu-Wing Tai$^1$\\
{\normalsize$^1$Tencent,\space\space\space\space $^2$Hong Kong University of Science and Technology}\\
{\tt\small yangzhaoyang6@126.com, zshiad@connect.ust.hk, goodshenxy@gmail.com, yuwingtai@tencent.com}
\end{tabular}
}

\maketitle

\begin{abstract}
Continuous sign language recognition (SLR) aims to translate a signing sequence into a sentence. It is very challenging as sign language is rich in vocabulary, while many among them contain similar gestures and motions. Moreover, it is weakly supervised as the alignment of signing glosses is not available. In this paper, we propose Structured Feature Network (SF-Net) to address these challenges by effectively learn multiple levels of semantic information in the data. The proposed SF-Net extracts features in a structured manner and gradually encodes information at the frame level, the gloss level and the sentence level into the feature representation. The proposed SF-Net can be trained end-to-end without the help of other models or pre-training. We tested the proposed SF-Net on two large scale public SLR datasets collected from different continuous SLR scenarios. Results show that the proposed SF-Net clearly outperforms previous sequence level supervision based methods in terms of both accuracy and adaptability.
\end{abstract}

\section{Introduction}

\begin{figure}[htb]
	\begin{center}
		\includegraphics[width=1\linewidth]{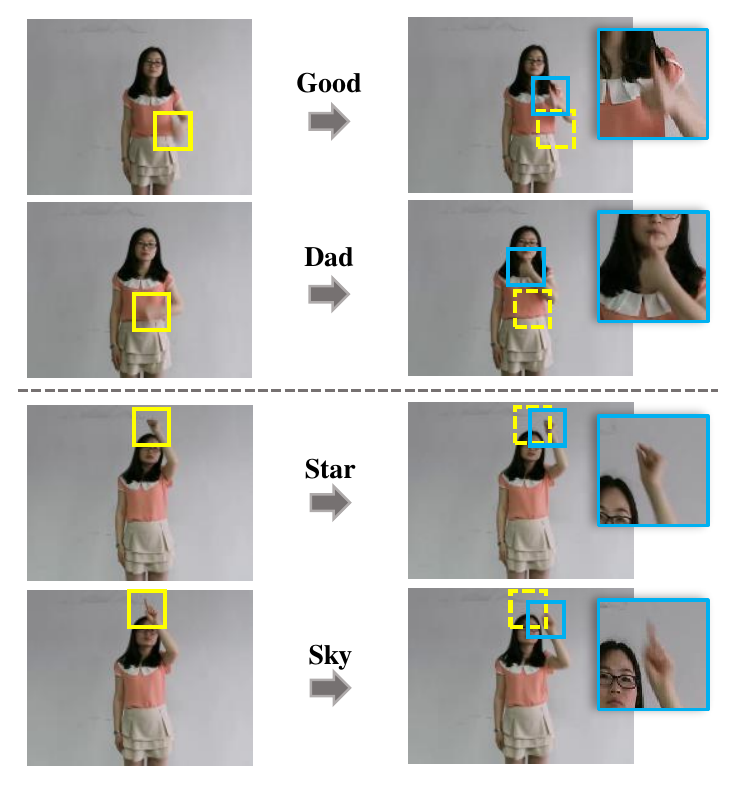}
	\end{center}
	\caption{Samples of glosses that look similar in the Chinese Sign Language Dataset \cite{lshan}. Yellow and blue boxes represent hand locations in the two frames. In each gloss pair, gloss on the top differs from the bottom one only either in motions or gestures. However, they are very different in meanings. }
    \label{fig:fig1}
\end{figure}

Sign language is considered to be the most structured form of gestural communication method. It is commonly used by deaf people as their major way of daily communications but is difficult for common people to understand. Gloss, which represents the closest meaning of a sign in the natural language, is generally defined to be the unit of the sign language \cite{gloss}. A gloss is typically made up by one or more hand gestures, motions, facial emotions and transitions in between them. A single change in one of these components can result in another sign that has a very different meaning (See Figure \ref{fig:fig1} for examples).

Continuous sign language recognition (SLR) aims to recognize glosses in a signing sequence. It is different from isolated SLR, in which each sign has been segmented and annotated independently. It is also different with sign language translation (SLT) \cite{nslt}, which involves an additional step to translate recognized glosses into a grammatical sentence. In continuous SLR, no segmentation and alignment but only the sentence level annotation for the whole signing sequence is given. This requires the model to learn not only frame level and gloss level features to distinguish different glosses, but also sentence level features to infer alignment and construct the sentence.

Recent years, deep learning \cite{deep} has achieved outstanding performance in many vision tasks. Successful work exists on applying deep learning techniques on continuous SLR \cite{deepsign,deephand,lshan,resign,cvpr19}. However, the task remains challenging even with deep learning. On the one hand, models used in these methods manage to learn features and alignments from the frame level. This could have limited the representativeness of features as single frames are far from the completion of a gloss. Also, the number of frames that a gloss lasts may vary dramatically, which could introduce uncertainty in alignment learning. On the other hand, some methods still need the help of additional models such as Hidden Markov Models (HMMs) or language models to construct the final sentence. This could have limited the adaptability of the method as it requires careful tweaking of the whole system for a specific dataset.

In order to address these challenges, in this paper, we propose Structured Feature Network (SF-Net). The proposed SF-Net learns features in a structured manner to gradually encode information at the frame level, the gloss level and the sentence level into the feature representation. The translated sentence can be obtained by doing greedy decoding using the final features. As a result, the alignment can be inferred from the gloss level rather than from the frame level. While different network designs are used for different levels of feature learning, the whole network can be trained end-to-end without the help of other models and pre-training.

We tested the proposed SF-Net on two large scale public SLR datasets, Chinese Sign Language (CSL) dataset \cite{lshan} and RWTH-PHOENIX-Weather-2014 dataset \cite{rwth}, which represent continuous SLR in a laboratory environment and real world respectively. The final results show that, the proposed SF-Net outperforms previous sequence level supervision based methods on both datasets. We also show in steps the effectiveness of several network designs in SF-Net.

\section{Related Work}

The work in this paper falls into the topic of sign language recognition (SLR), which is also related to sequence to sequence learning and human action and gesture recognition. We hereby provide a literature review on these topics.

{\bf Sign Language Recognition. }SLR can be divided into isolated SLR and continuous SLR. Isolated SLR discusses scenarios where signs are segmented so that each sample of data contains only one running gloss as the recognition target. Much work exists on successfully recognizing isolated signs \cite{iso1,iso2,iso3}. Differently, continuous SLR is a more challenging scenario where several running glosses are signed in a continuous sequence in a single sample of data. In this case, the task is weakly supervised as signs are not segmented, and only the overall transcripts are given, without temporal alignment information. Most existing continuous SLR methods divide the task into three stages, including temporal segmentation or alignment learning, isolated SLR, and sentence construction with language models \cite{dtwhmm,resign}. While these methods have achieved convincing performance, they may have to be trained with additional supervisions or the help of other pre-trained models, which requires careful tweaking of the whole system for a specific dataset. End-to-end continuous SLR methods also exist \cite{subu,rec,lshan}. However, these methods learn features and alignments on the frame level, which may fail to fully investigate the semantic information in the data. Differently, the proposed SF-Net can capture different levels of information in the data by extracting features in a structured manner. Another related topic in the scope would be sign language translation (SLT) \cite{nslt}, which takes SLR as the first step and adds one additional step to translate recognized glosses into common sentences. In this paper, we focus on the SLR part only. 

{\bf Sequence to Sequence Learning. }It is natural to think of continuous SLR as a sequence to sequence task as it translates a sequence of running glosses into a sequence of words. Most methods in this topic are Encoder-Decoder framework \cite{ed} Connectionist Temporal Classification (CTC) based. The Encoder-Decoder framework nowadays generally incorporates the attention mechanism \cite{att} to learn long term dependencies and the alignment between the source sequence and the target sequence. Alternatively, CTC aims to learn a comprehensive scoring function for the whole sequence instead of classification scores for each of the single frames. These two methods have been successfully applied to speech recognition \cite{deepsp2,spstate}, text recognition \cite{ctcw,attw}, video captioning \cite{cap1,cap2} and neural machine translation \cite{trans}. However, the unit in these tasks can be easily defined and processed (for example, source words in language translation). This is different from continuous SLR as the unit, which is supposed to be gloss, can hardly be pre-defined as they vary a lot in length. Moreover, Encoder-Decoder framework based methods generally need a lot of data to learn the mapping between the source and the target sequence. This is not available for continuous SLR.

{\bf Action and Gesture Recognition. }Continuous SLR shares some similarities with action and gesture recognition as they all discuss body language. However, they are mostly based on different foundations. Gesture recognition generally discuss stationary hand shape or body postures. Most efforts fall on detecting key part of the body (such as hands) which may have significant impact on the following classification \cite{ges,3dhand}. Action recognition is closer to SLR as they both learn body motions in time series. Recent advances in network architecture have considerably improved the performance of action recognition in benchmark datasets \cite{slowfast,ac,mict}. However, actions in each of the samples are complete and well-defined, making them suitable for classification. On the contrary, several different glosses may appear in a continuous sequence in continuous SLR. Nonetheless, network architectures that can extract action features effectively have given us insights in designing the proposed SF-Net.

\section{Structured Feature Network}

\begin{figure*}[htb]
	\begin{center}
		\includegraphics[width=1\linewidth]{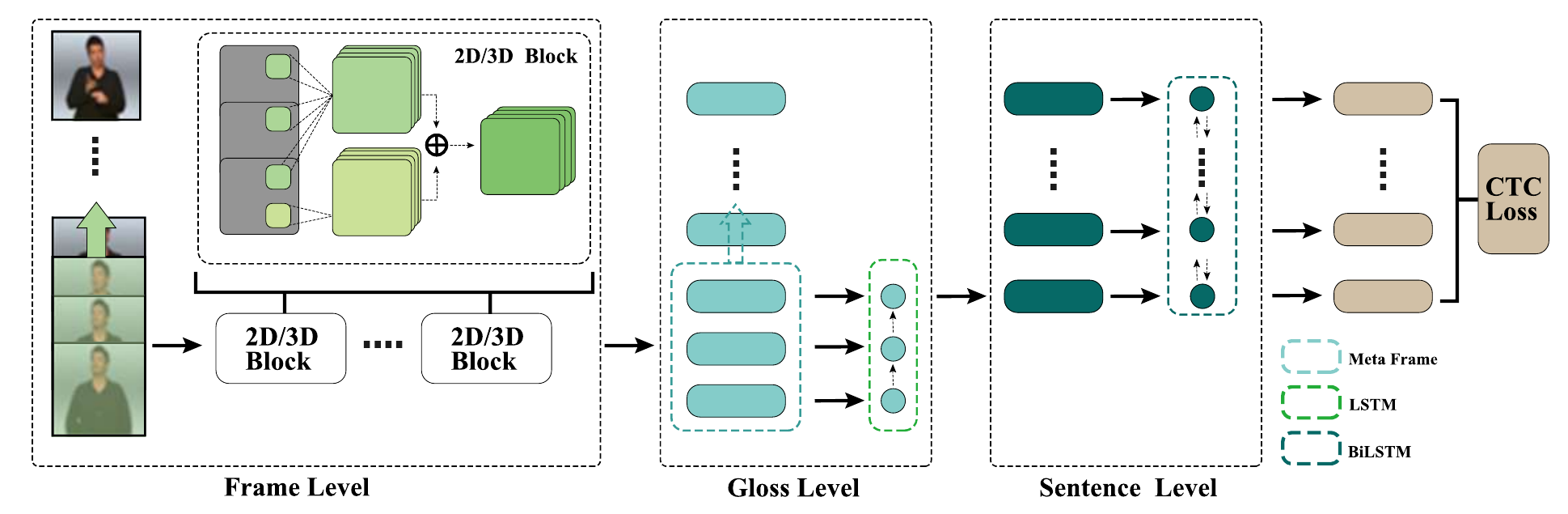}
	\end{center}
	\caption{Overview of the proposed SF-Net. Squares in the figure are feature maps while strip-shape rectangles are one-dimension feature vectors. Their copies represent the expansion in the temporal dimension. The three levels of feature extraction are distinguished using different colors. }
    \label{fig:fig2}
\end{figure*}

Continuous SLR takes a sequence of signing frames as input and learn to directly output the target sequence of glosses in the right order. In this task, there are implicitly three levels of information in the data that need to be considered. First, the frame level. The signing gesture and facial emotion are important information for distinguishing different glosses. They are the bottom most level of information in the task and can be captured by processing and extracting features in the frames. Second, the gloss level. Signing glosses are made up of several gestures, emotions and motions (in fact, we can also consider holding on a gesture as one kind of body motion). As a result, independent frames are far from the completion of a gloss. Therefore, information of several frames has to be combined to form features in this level. Finally, the sentence level. Different glosses are performed in a continuous sequence without explicit segmentation. In order to align and translate the signing sequence to target sentence, gloss level features need to be re-organized in this level so that context information in the sequence can be encoded.

We propose the Structured Feature Network (SF-Net). Unlike previous methods that may not have fully investigated the information discussed above, the proposed SF-Net uses different network designs to learn features in three levels which can be paired to the levels of information in the task. By effectively learning features in this structured manner, information at the three levels can be gradually encoded into the final feature representation and the task can be made end-to-end trainable without the help of other methods or pre-training. An overview of the proposed SF-Net is shown in Figure \ref{fig:fig2}.

\subsection{Frame Level Feature}

Feature learning at this level focuses on the gestures and facial emotions in each frame. Like in many other applications, this can be effectively done by stacking up several 2D convolutional layers. As each sample in continuous SLR is a sequence of signing frames, a mini-batch of samples can be represented as a 5D matrix $I^{B \times T \times C \times H \times W}$, where $B$, $T$, $C$, $H$, $W$ denotes the batch size, the length of the sequence, the number of channels in each frame, and the height and width of each frame respectively. The 2D convolution can be then done per sample per frame as:
\begin{equation}
{Y}_{i,t,k,y,z}^{2D} = \sum_{c=0}^{C-1}\sum_{h=0}^{H-1}\sum_{w=0}^{W-1} I_{i,t,c,h+y,w+z}K_{k,c}^{2D}
\label{eqn:equa1}
\end{equation}
where $i$, $t$, $k$, $y$, $z$ are indexes of the output, ${Y}$ the output and $K^{2D}$ the 2D kernel.

This operation treats each frame independently, which may have some shortcomings in extracting features for sign language. This is because that there are many fast and small motions (such as quick finger movements) in sign language. These motions last only for a few frames and the difference between these frames may be too small to observe without comparing them directly. Therefore, in order to capture these fast and small motions, we propose to incorporate 3D convolutional layers \cite{3dcnn} that take adjacent frames into account during feature extraction in the frame level. The 3D convolutional is done per sample as:
\begin{equation}
{Y}_{i,x,k,y,z}^{3D} = \sum_{c=0}^{C-1}\sum_{t=0}^{T-1}\sum_{h=0}^{H-1}\sum_{w=0}^{W-1} I_{i,t+x,c,h+y,w+z}K_{k,c}^{3D}
\label{eqn:equa2}
\end{equation}
where $K^{3D}$ is the 3D kernel. We do not reduce the temporal dimension during 3D convolution, so in SF-Net $X \equiv T$.

Inspired by the MiCT Network \cite{mict}, after each 2D and 3D convolution, we merge features of the two branches with an cross domain element-wise summation. As a result, the final output is:
\begin{equation}
{Y}_{i,t,k,y,z} = {Y}_{i,t,k,y,z}^{3D} + {Y}_{i,t,k,y,z}^{2D}
\label{eqn:equa3}
\end{equation}

This operation can speed up learning and allow training of deeper architectures. At the same time, it allows the 3D convolution branch only to learn residual temporal features, which is the fast and small motions in sign language for us, to compensate features learned in 2D convolution. These 3D convolutions have actually added another sub-level of feature learning in the frame level. As a result, instead of stacking up 2D convolution layers, we use several 2D/3D convolution blocks in the frame level of feature extraction (as shown in Figure \ref{fig:fig2}). After the last convolution block, we conduct a global average pooling to reduce dimension. The final feature will be of dimension $Y^{B \times T \times K}$, where $K$ is the number of channels in the last block.

\subsection{Gloss Level Feature}

\begin{figure}[htb]
	\begin{center}
		\includegraphics[width=1\linewidth]{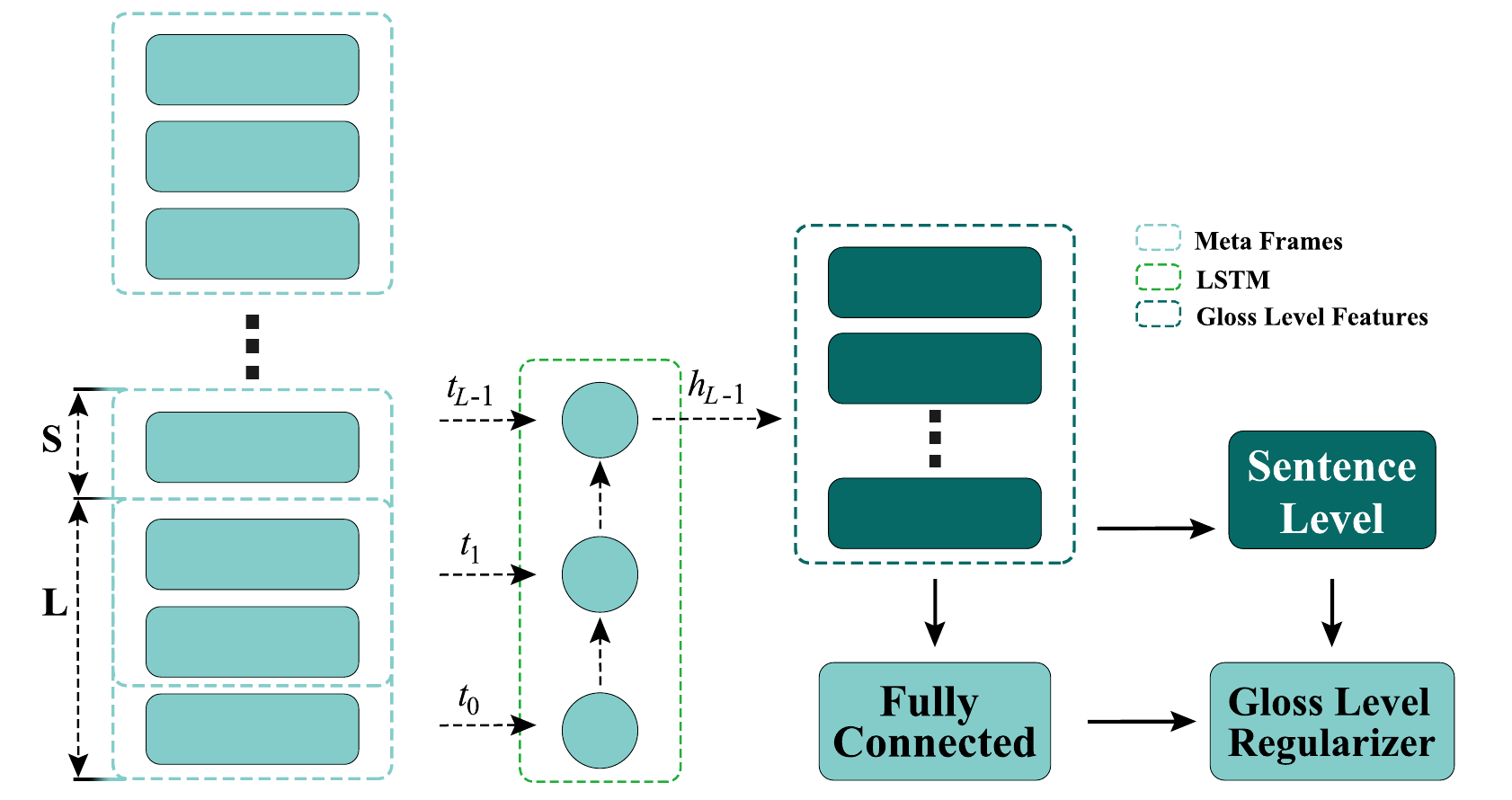}
	\end{center}
	\caption{Network design of the gloss level part of SF-Net. A framing operation is added in this level to capture gloss level motions and a regularizer is introduced to prevent overfitting. Framing settings in the figure have a window size of 3 and stride of 1. These settings are for illustration purpose only. }
    \label{fig:fig3}
\end{figure}

Gloss is the unit of the sign language. However, in continuous SLR, the segmentation of these units is not available. This requires the network to align certain frames to a corresponding gloss in the target sentence. This alignment is hard to learn as isolated frames are far from the completion of a gloss. Although features in the frame level have encoded some fast motion information, the number of frames considered are still much smaller than the number of frames a gloss can last. It is therefore necessary to add a new level of feature learning to better encode gloss level information. We show the network design in this level in Figure \ref{fig:fig3}.

Inspired by the framing step in automatic speech recognition (ASR), we also add a framing step after the frame level feature extraction. Similar to the framing in ASR, given an input of length $T$, the window size $L$ and the stride $S$, the number of meta frames generated is:
\vspace{-0.05in}
\begin{equation}
F = \floor{\frac{T-L}{S}} + 1
\label{eqn:equa4}
\vspace{-0.05in}
\end{equation}
and each meta frame contains $L$ frames and is of dimension $[L \times K]$. After framing, the output of the frame level feature will be transformed into a 4D matrix of dimension $Y'^{B \times F \times L \times K}$.

In order to reduce dimension and form more compact gloss level features, we add a long short term memory (LSTM) layer to learn the temporal dependencies between frames in meta frames. The LSTM layer can encode temporal information into the feature while also preserve the ordering of frames during encoding. This is an important reason that we choose LSTM over the others as the signing ordering is also key to distinguish glosses. By taking out the hidden state of the last frame as the feature of each meta frame, the output dimension of features in this level becomes $M^{B \times F \times H}$, where $H$ is the number of hidden nodes in the LSTM layer.

Note that the combination of the LSTM and the 3D convolution in the frame level has actually created an effective temporal modeling architecture, where the 3D convolution takes care of the short term fast motions and the LSTM learns slower motions that have longer temporal dependencies. This has fitted in the pattern of sign language as both slow and fast motions can appear in signing a gloss.

Furthermore, as the number of data available for training continuous SLR is limited, to prevent overfitting and fully develop the network capacity in the first two levels, we added a regularizer in the gloss level to enhance the generalization of the features. We first used a fully-connected layer and a followed Softmax activation to transform features of meta frames $M^{B \times F \times H}$ into a probability distribution, where each entry in the distribution represents the likelihood of the meta frame being the corresponding gloss in the vocabulary.  Then, the regularizer is realized without additional supervision by forcing these distributions to be close to the ones obtained in the sentence level. Specifically, let $P^{gl}$ be the probability distribution obtained in the gloss level and $P^{sl}$ the one obtained in the sentence level (which is also the one used for emitting the final output), we use {Kullback}-{Leibler} Divergence Loss:
\vspace{-0.05in}
\begin{equation}
L_{g} = -\sum_{n=1}^{N} P^{sl}_nlog(\frac{P^{gl}_n}{P^{sl}_n})
\label{eqn:equa5}
\vspace{-0.05in}
\end{equation}
where $N$ is the vocabulary size.

This regularizer is introduced after the first few epochs to ensure stable training.

\subsection{Sentence Level Feature}

Context information is important for continuous SLR and other sequence to sequence tasks to learn the alignment between the source and the target sequence. In this last level of sentence feature learning, we follow a standard setup used in many other sequence to sequence tasks. We add a Bi-Directional LSTM (BiLSTM) which takes as input the gloss level feature $M^{B \times F \times H}$ and re-organize these features to encode context information in both directions into the feature representation. The final sentence level feature will be of dimension $O^{B \times F \times 2H}$, as features in the two directions will be concatenated.

These features are then fed into a fully connected layer that casts them into the prediction space. We choose the Connectionist Temporal Classification (CTC) \cite{ctc} as the loss function over the Encoder-Decoder framework as it tends to get overfitting in seen target sequence patterns. As a result, the loss function is:
\vspace{-0.05in}
\begin{equation}
L_{s} = L_{CTC} = -log({P(\pmb{y}|\pmb{x})})
\label{eqn:equa6}
\vspace{-0.05in}
\end{equation}
where $\pmb{y}$ is the target sequence of glosses and ${P(\pmb{y}|\pmb{x})}$ is the sum of probabilities of all decoding paths that will result in $\pmb{y}$ after collapsing repetitions and removing blanks.

When combined with the regularizer in the gloss level, the loss function becomes:
\vspace{-0.05in}
\begin{equation}
L = L_{s} + [E > E_{start}] \cdot L_{g}
\label{eqn:equa7}
\vspace{-0.05in}
\end{equation}
where $E$ is the current training epoch index and $E_{start}$ is the epoch index that the regularizer will be introduced. During testing, the final output can be obtained by simply doing greedy decoding on the probability $P^{sl}$.

\section{Experiments}

We conducted experiments on two large scale continuous SLR datasets, Chinese Sign Language (CSL) dataset \cite{lshan} and RWTH-PHOENIX-Weather-2014 dataset \cite{rwth}. We show evidence on the effectiveness of several design choices of the proposed SF-Net and also compare it to other methods. Qualitative results of SF-Net on full videos are provided in the Appendix.

\subsection{Datasets}

Chinese Sign Language (CSL) dataset \cite{lshan} is a dataset collected in a laboratory environment. There are 50 signers and 100 unique sentences in the dataset. Each signer has performed each of the 100 sentences for 5 times, giving in total 25,000 samples and more than 100 hours footage. Videos are collected with a Microsoft Kinect camera and post-processed to a unified resolution of 720 $\times$ 1280 and frames per second (FPS) of 30. The dataset also has a word-level version, where the same 50 signers have each performed 500 unique words once. As no official split is provided, we did the split ourselves and gave 20,000 and 5,000 samples to the training set and testing set respectively. When splitting the dataset, we have ensured that signers have no overlap in the two sets.

RWTH-PHOENIX-Weather-2014 dataset \cite{rwth} is a real world SLR dataset which represents a more challenging scenario. It is recorded from a public television broadcast in Germany. It contains 6841 unique sentences performed by 9 signers. Signers all wear dark clothes and sentences are performed in front of an artificial grey background. There are about 80,000 running glosses in the dataset, giving in total more than 10 hours in length. It is much richer in vocabulary compared to the CSL dataset, which is of size 1231. Videos have been post-processed to a unified resolution of 210 $\times$ 260 and an FPS of 25. We follow the official split of the dataset, which gives 5672, 540, 629 samples to training, validation and testing respectively.

\subsection{Settings}

\begin{figure}[htb]
	\vspace{-0.15in}
	\begin{center}
		\includegraphics[width=0.93\linewidth]{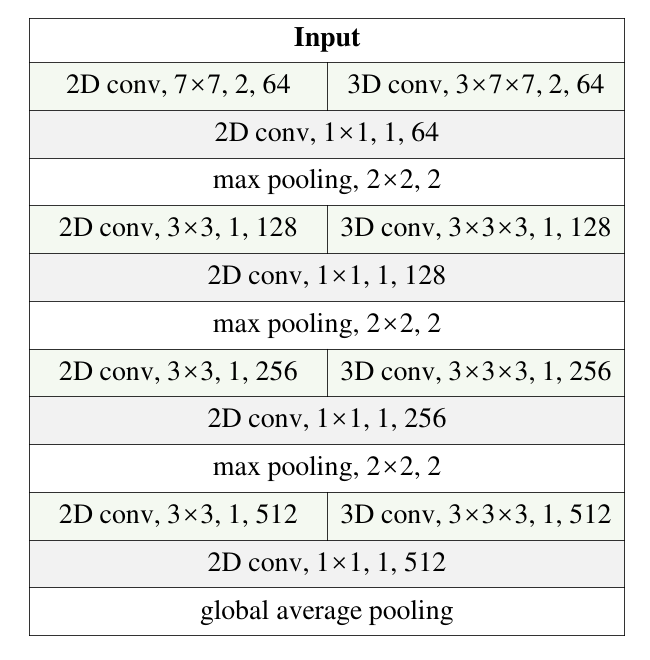}
	\end{center}
	\vspace{-0.32in}
	\caption{Network settings for the frame level part. }
    \label{fig:fig6}
	\vspace{-0.05in}
\end{figure}

Our settings for the frame level part is shown in Figure \ref{fig:fig6}. In the rest of the network, we use 1 LSTM layer with 512 hidden nodes and 1 BiLSTM layer with 256 hidden nodes in each direction respectively for the gloss level and sentence level part of the network. The window size $L$ we choose for the gloss level framing is 12, which is approximately 0.5 seconds for both datasets. The framing stride $S$ is set to 3. Batch normalization \cite{bn} is used after every 2D convolution and 2D/3D blocks. Moreover, sequence-wise batch normalization \cite{deepsp2} is used for LSTM and BiLSTM layers.

For the CSL dataset, we central cropped all video frames to reduce the blank area in the frames. We then resized the frames to 224 $\times$ 224 as a final step of pre-processing. For the RWTH-PHOENIX-Weather-2014 dataset, we simply resized all frames to 256 $\times$ 256 and random cropped a 224 $\times$ 224 area as a way of data augmentation during training. We used Adam optimizer \cite{adam} for training the networks with an initial learning rate of 1e-4 and a weight decay of 1e-5. Learning rate was decreased by a factor of 0.5 in the half way of training. We trained the network for 40 and 60 epochs respectively for the two datasets.

We use the word error rate (WER) as the evaluation metric for the purpose of comparison with results reported in other work. It is defined as:
\begin{equation}
WER = \frac{\# substitution + \# deletion + \# insertion}{\# words\ in\ the\ target}
\label{eqn:equa8}
\end{equation}

Note that, when the output is Chinese, we consider each Chinese character as a unique word for better comparison with results of previous methods.

\subsection{Network Design Analyses}

\begin{figure*}[htb]
	\begin{center}
		\includegraphics[width=1\linewidth]{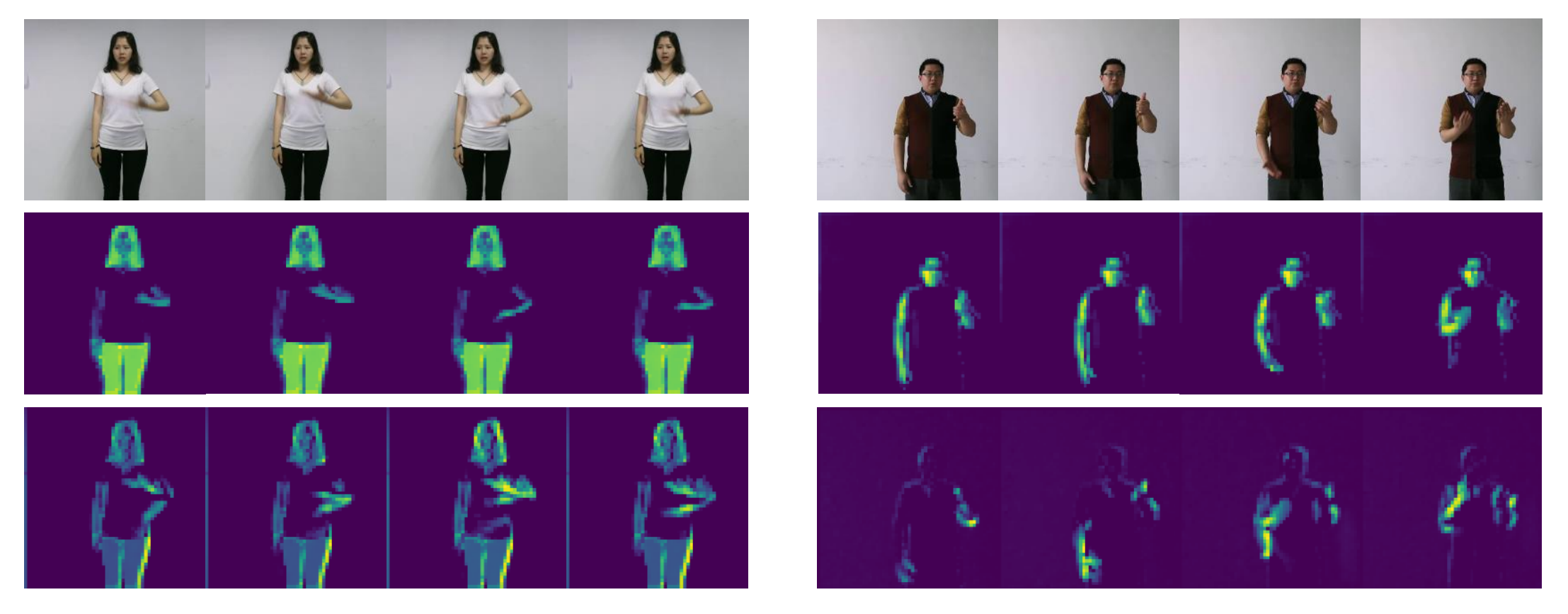}
	\end{center}
	\caption{Comparison of feature maps after the first convolution (or block). We show 4 frames of 2 samples downsampled from 16 frames in the original sequence. For each sample, the first row is the original frame, the second row is the feature map learned without 3D convolution and the last row is the feature map learned with 3D convolution. In each of the feature maps, areas that have been given more attention are colored brighter. }
    \label{fig:fig4}
	\vspace{-0.2in}
\end{figure*}

{\bf 2D/3D Convolution Block. }We first tested the effectiveness of adding additional branches of 3D convolutions in the frame level feature extraction. We conducted experiments on both the word-level CSL dataset and sentence-level CSL dataset and compared the performance of the SF-Net when training with and without 3D convolutions. When doing experiments on the word-level dataset, the sentence level part of SF-Net has been removed. Also, a fixed length of 2s (60 frames) of video is cut out from a random position in the original word-level videos and downsampled to contain only 12 frames. By doing this, the framing stage in the gloss level part of SF-Net would only generate one meta frame. We use the feature of this meta frame for classifying the video. Results on the testing set are shown in Table \ref{table:tab1}.

We can see that, the 3D convolution branch has brought nearly 3.5\% of accuracy gain compared to the accuracy obtained without 3D convolution in the word-level classification. Similarly, in sentence-level, the WER has reduced for more than 2\%. This indicates that the fast and small motions that exist in sign language are indeed important information for distinguishing glosses. This information can be successfully captured by 3D convolutions. We give a feature map comparison in Figure \ref{fig:fig4} for further analyses.

\begin{table}[!htpb]
 \centering
 \fontsize{3.5}{3.5}\selectfont
 \vskip 0.05in
 \renewcommand{\arraystretch}{1.3}
 \setlength\arrayrulewidth{0.15pt}
 \resizebox{0.95\linewidth}{!}{
  \begin{tabular}[c]{c|ccc}
    \hline
    & {\bf Word} & {\bf Sentence}\\ \hline
   {\bf Without 3D} & 17.3  & 7.1 \\
   {\bf With 3D} & \bf 13.0 & \bf 4.7 \\ \hline
  \end{tabular}
 }
\caption{Comparison of performance when training with and without 3D convolutions. Results are classification error rate and WER for the word-level and sentence-level respectively. }
\label{table:tab1}
\end{table}

It can be observed that, feature maps learned by 2D convolutions simply have highlights at arm, head and leg positions in the current frame. On the contrary, after 3D convolution is introduced, feature maps transformed to either have additional highlights at arm or hand positions in adjacent frames, or only have highlights for the moved portion of the body. Both can be a way of encoding fast motions. Moreover, we can see that feature maps learned by 2D/3D convolution block have shown fewer highlights in irrelevant areas, such as at leg areas. This may own to the branch merging strategy which helps achieve a better gradient propagation. Both these properties of 2D/3D convolution block can help stabilize learning and improve the final performance.

\begin{figure*}[htb]
	\begin{center}
		\includegraphics[width=1\linewidth]{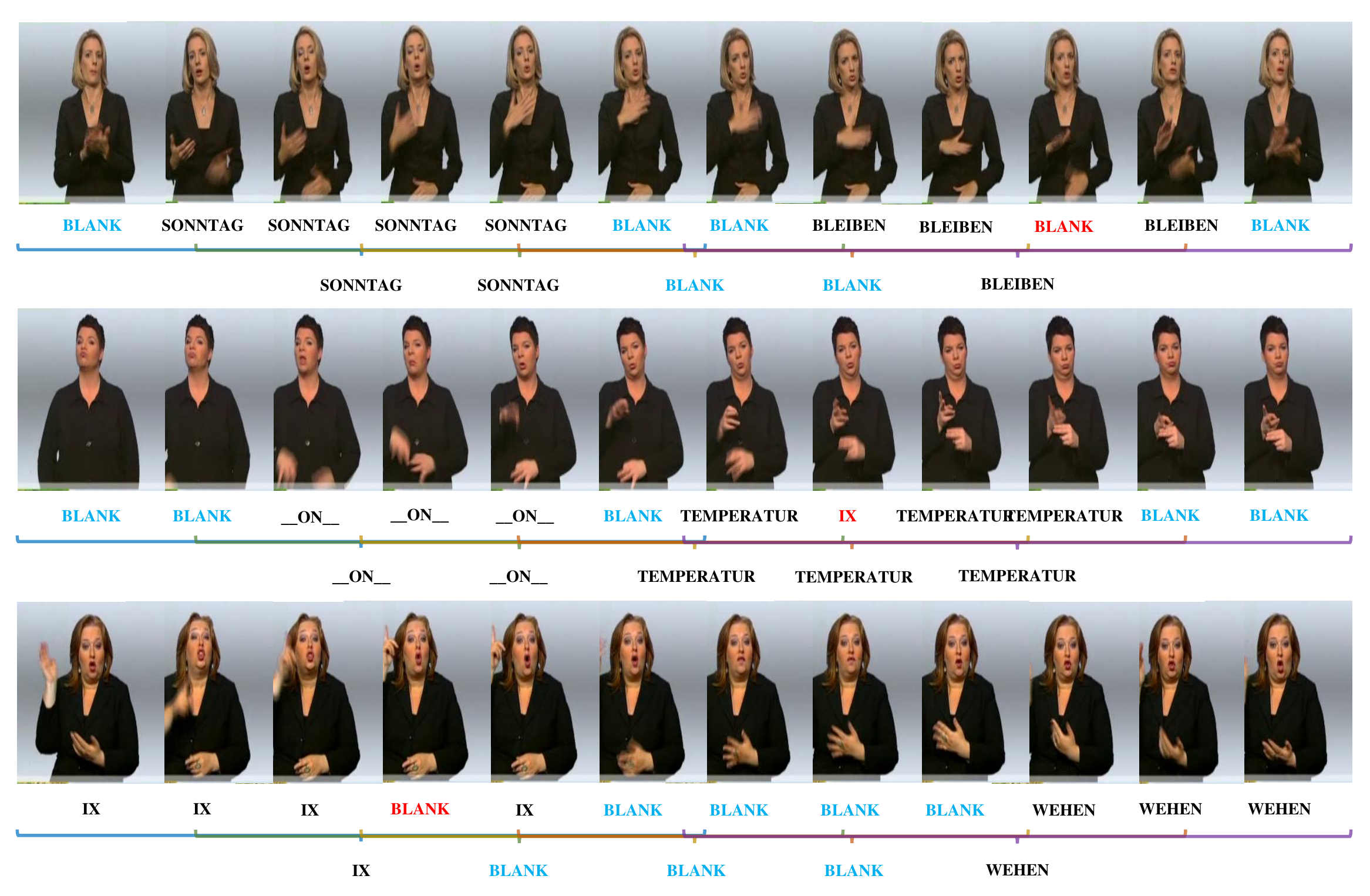}
	\end{center}
	\caption{Comparison of alignment. We show 12 frames of 3 samples downsampled from 24 frames in the original sequence. For each sample, we first show outputs given by SF-Net trained without framing, and then the ones of SF-Net trained with framing and LSTM. Different bracket colors indicates different meta frames. Blank outputs are colored in blue, which will be removed in decoding. Outputs that will cause errors in decoding are colored in red. }
    \label{fig:fig5}
	\vspace{-0.17in}
\end{figure*}

{\bf Gloss Level Feature. }We then tested the effectiveness of adding the gloss level feature extraction. We conducted experiments on both the CSL dataset (both word-level and sentence level) and the RWTH-PHOENIX-Weather-2014 dataset and compared the performance of the SF-Net when training with and without the gloss level part. After removing the gloss level part, we tried two approaches: do the framing but simple concatenate features in each meta frame without going through the LSTM layer, and remove both framing and LSTM where output features from the frame level will be fed into the sentence level directly. For word-level CSL dataset, only the former approach is used. Gloss level regularizer was not used in this set of experiments. Results on the testing set are shown in Table \ref{table:tab2}.

\begin{table}[!htpb]
 \centering
 \small
 \vskip 0.05in
 \renewcommand{\arraystretch}{1.3}
 \setlength\doublerulesep{1pt}
 \resizebox{0.95\linewidth}{!}{
  \begin{tabular}[c]{c|cccc}
    \hline
    & \multicolumn{2}{c}{\bf CSL} & \multirow{2}{*}{\bf RWTH}\\ \cline{2-3}
    & {\bf Word} & {\bf Sentence}\\ \hline
   {\bf Without framing} & -  & 11.9 & 46.7\\
   {\bf With framing} & 19.1  & 8.8 & 45.0\\
   {\bf With LSTM} & \bf 13.0 & \bf 4.7 & \bf 40.8\\ \hline
  \end{tabular}
 }
\caption{Comparison of performance when training with and without gloss level feature extraction. Results are classification error rate for word-level CSL dataset and WER for sentence-level CSL and RWTH datasets. }
\label{table:tab2}
\vspace{-0.1in}
\end{table}

We can see that, there is a dramatic drop of performance for both datasets when the framing and LSTM in the gloss level part of the SF-Net are removed. This may mainly because that inferring the alignment between the input and the output directly from the frame level is much harder as supervision is only given on the whole sentence per glosses but not per frame states. Without framing, the searching space in decoding greatly increased and this may require more powerful context information encoding to learn.

We can also see that, when framing is used with the absence of LSTM layer, the results get better but still far below the performance when LSTM is used. This indicates that modeling of the temporal information in each meta frame is also important. Otherwise, there are may be too many redundant information to achieve effective sentence level learning. Finally, by fully implementing the gloss level design of the SF-Net, we achieved the best results in this set of experiments. We show three alignment samples in Figure \ref{fig:fig5} to better reveal how framing has improved performance.

The errors made by the model that is trained without framing are typical types of errors that we observed in frame level alignment prediction, where one error can distort the whole sequence output. On the contrary, this has been alleviated after framing is introduced. Framing has made the output prediction become much sparser (24 predictions (only show 12 due to page limit) compared to 5 predictions in Figure \ref{fig:fig5}), which can reduce the probability of introducing these errors. Furthermore, the prediction becomes more accurate as the LSTM has encoded the temporal dependencies between frames in each meta frame into the feature representation.

{\bf Gloss Level Regularizer. }Finally, we tested the effectiveness of having an addition regularizer in the gloss level. We conducted comparison experiments on the RWTH-PHOENIX-Weather-2014 dataset. We also tuned the value of $E_{start}$ to see the impact of adding the regularizer in different stages of training. Results are shown in Table \ref{table:tab3}.

\begin{table}[!htpb]
 \centering
 \small
 \vskip 0.05in
 \renewcommand{\arraystretch}{1.3}
 \setlength\doublerulesep{1pt}
  \begin{tabular}[c]{c|ccccc}
    \hline
    & \multirow{2}{*}{\bf No reg} & \multicolumn{4}{c}{\bf Epoch}\\ \cline{3-6}
    &  & \bf 1 & \bf 5 & \bf 15 & \bf 25\\ \hline
   {\bf WER} & 40.8 & 42.7 & 40.2 & 38.4 &\bf 38.1 \\ \hline
  \end{tabular}
\caption{Comparison of performance on the RWTH-PHOENIX-Weather-2014 dataset when adding gloss level regularizer in different stages of training. }
\label{table:tab3}
\vspace{-0.15in}
\end{table}

It can be observed that, when the regularizer is introduced in the early stage of training, the performance has dropped for nearly 2\%. This may be because that the network can have very unstable output probability distributions in the early stage of training. This has made learning difficult and resulted in worse convergence. Different, when we add the regularizer in the medium stage of training, it helped improve the final performance for more than 2.5\%. This has demonstrated its effectiveness.

However, we did not find similar observations when training the CSL dataset and it seems that the regularizer has little impact on the result. We believe it is because that the vocabulary size, sentence length and possible combinations of glosses are smaller in the CSL dataset. On the contrary, the RWTH-PHOENIX-Weather-2014 dataset has a richer vocabulary and contains non-repeated, longer sentences, and some glosses appear only for a few times. When learning on it, regularizer can help to prevent overfitting on seen sentence patterns and better develop the capacity of the first two levels of the network.

\subsection{Overall Performance}

We did a thorough comparison between the performance of the proposed SF-Net and previous methods. To fully investigate the performance of the proposed SF-Net, we added a set of experiments where we initialized the frame level and the gloss level parts of the network with parameters learned in training the word-level CSL dataset. This can help accelerate learning, though we observed that similar results can be obtained by training from scratch after adding the number of training epoch. Moreover, to fully investigate the capacity of the algorithm, we also conducted a set of experiments where we used ResNet-18 \cite{res} as our backbone architecture with all non-bottleneck layers changed to 2D/3D convolution blocks. For fair comparison, we only considered previous methods that are based on sentence level supervision excluding methods using frame-level labels such as \cite{resign}). Methods that use other kinds of supervision (such as frame state labels) are not included in this section. We report the result in Table \ref{table:tab4} and Table \ref{table:tab5} for the two datasets respectively. Most results for other methods are collected from their original papers or dataset release papers. We only re-trained SubUNet \cite{subu} for the CSL dataset.

\begin{table}[!htpb]
 \centering
 \fontsize{2}{2}\selectfont
 \vskip 0.05in
 \renewcommand{\arraystretch}{1.3}
 \setlength\arrayrulewidth{0.1pt}
 \resizebox{0.95\linewidth}{!}{
  \begin{tabular}[c]{c|c}
    \hline
   {\bf Methods} & {\bf WER}\\ \hline
   {\bf DTW-HMM \cite{dtwhmm}} & 28.4 \\
   {\bf LSTM \cite{lstm1}} & 26.4 \\
   {\bf S2VT \cite{s2vt}} & 25.5 \\
   {\bf LSTM-A \cite{lstma}} & 24.3 \\
   {\bf LSTM-E \cite{lstme}} & 23.2 \\
   {\bf HAN \cite{han}} & 20.7 \\
   {\bf LS-HAN \cite{lshan}} & 17.3 \\
   {\bf SubUNet \cite{subu}} & 11.0 \\ \hline
   {\bf SF-Net (scratch)} & \bf 4.8 \\
   {\bf SF-Net} & \bf 3.8 \\ \hline
  \end{tabular}
 }
\caption{Comparison of performance of different methods on the CSL dataset. }
\label{table:tab4}
\vspace{-0.1in}
\end{table}

\begin{table}[!htpb]
 \centering
 \fontsize{3.2pt}{3.2pt}\selectfont
 \vskip 0.05in
 \renewcommand{\arraystretch}{1.3}
 \setlength\arrayrulewidth{0.2pt}
 \resizebox{0.95\linewidth}{!}{
  \begin{tabular}[c]{c|cc}
    \hline
   \multirow{2}{*}{\bf Methods} & \multicolumn{2}{c}{\bf WER}\\ \cline{2-3}
    & {\bf Dev} & {\bf Test}\\ \hline
   {\bf \cite{rwth}} & 57.3 & 55.6 \\
   {\bf Deep Hand \cite{deephand}} & 47.1 & 45.1 \\
   {\bf Deep Sign \cite{deepsign}} & 38.3 & 38.8 \\
   {\bf SubUNet \cite{subu}} & 40.8 & 40.7 \\
   {\bf \cite{rec}} & 39.4 & 38.7 \\
   {\bf LS-HAN \cite{lshan}} & - & 38.3 \\
   {\bf Align-iOpt \cite{cvpr19}} & 37.1 & 36.7 \\ \hline
   {\bf SF-Net (scratch)} & \bf 38.0 & \bf 38.1 \\
   {\bf SF-Net} & \bf 36.5 & \bf 36.1 \\
   {\bf SF-Net(ResNet-18)} & \bf 35.6 & \bf 34.9 \\ \hline
  \end{tabular}
}
\caption{Comparison of performance of different methods on the RWTH-PHOENIX-Weather-2014 dataset. }
\label{table:tab5}
\vspace{-0.2in}
\end{table}

We can see that the proposed SF-Net has achieved the best performance among these methods on both two datasets, even when training from scratch. When training from pre-learned parameters in the word-level CSL dataset, we observed further improvements in accuracy. This has demonstrated the effectiveness and adaptability of SF-Net on learning in different scenarios.

However, we should still note that the performance on the RWTH-PHOENIX-Weather-2014 dataset is far from satisfactory for real world applications. The high diversity, large vocabulary, limited number of training samples and the weakly supervised nature of the task are all factors that have made this dataset challenging. Adding more regularizer or data are possible future work directions to level up the performance. Moreover, sign language is highly regional due to the lack of spreading, educating and standardizing, which has ended up with the co-existence of many different variations of the language around the world. This has dragged behind the development of algorithm and larger dataset in SLR. More work has to be done in bridging this gap in the future.

\section{Conclusions}

In this paper, we propose Structured Feature Network (SF-Net) to extract features from three levels of information that co-exist in continuous SLR. In the frame level, the proposed SF-Net incorporates 2D and 3D convolution to capture gesture, emotion and fast and small motion information. Then a framing step is added in the gloss level to generate meta frames which will be processed by LSTM to form gloss level features. These features will be further re-organized by the BiLSTM in the sentence level to encode context information.

We tested the proposed SF-Net on the CSL and the RWTH-PHOENIX-Weather-2014 datasets. Results have demonstrated the effectiveness of several designs in the network. Results also show that the proposed SF-Net has outperformed previous sentence level supervision based methods, in terms of both accuracy and adaptability.

{\small
\bibliographystyle{ieee}
\bibliography{egbib}
}

\begin{appendix}
\renewcommand{\thesection}{\Alph{section}}
\renewcommand{\thefigure}{S\arabic{figure}}
\setcounter{figure}{0}
\renewcommand{\thetable}{S\arabic{table}}
\setcounter{table}{0}
\section{Appendix}
\subsection{Framing Window Size}

We conducted a set of experiments on the RWTH-PHOENIX-Weather-2014 dataset \cite{rwth} to investigate the impact of the framing window size on the final performance. We fully implemented the proposed SF-Net without the frame level regularizer and only tuned the window size. Results are in Table \ref{table:tab2s}.

\begin{table}[!htpb]
 \centering
 \normalsize
 \vskip 0.05in
 \renewcommand{\arraystretch}{1.3}
 \setlength\doublerulesep{1pt}
 \resizebox{0.95\linewidth}{!}{
  \begin{tabular}[c]{c|ccccccc}
    \hline
    & \multirow{2}{*}{\bf No frame} & \multicolumn{6}{c}{\bf Window Size}\\ \cline{3-8}
    &  & \bf 3 & \bf 6 & \bf 9 & \bf 12 & \bf 15 & \bf 18 \\ \hline
   {\bf WER} & 46.7 & 45.9 & 43.1 & \bf 40.7 & 40.8 & 41.0 & 41.5 \\ \hline
  \end{tabular}
 }
\caption{Comparison of performance on the RWTH-PHOENIX-Weather-2014 dataset when using different framing window sizes. }
\label{table:tab2s}
\end{table}

We can see that the performance has dropped when the window size is very small (3 or 6 frames). This may be because the number of frames is too small to really learn gloss level temporal dependencies, as we observe most glosses take around 500 ms to perform. Then, the performance stays relatively stable for window size from 9 to 18, even we can observe a tendency of performance decline if the size continue to grow. However we were not able to further increase it as we have to make the output sequence length to be longer than the target sequence length. We set the window size at 12 as to maximally reduce the number of meta frames without hurting the performance.

\subsection{Qualitative Results}

\begin{figure*}[htb]
    \begin{center}
        \includegraphics[width=1\linewidth]{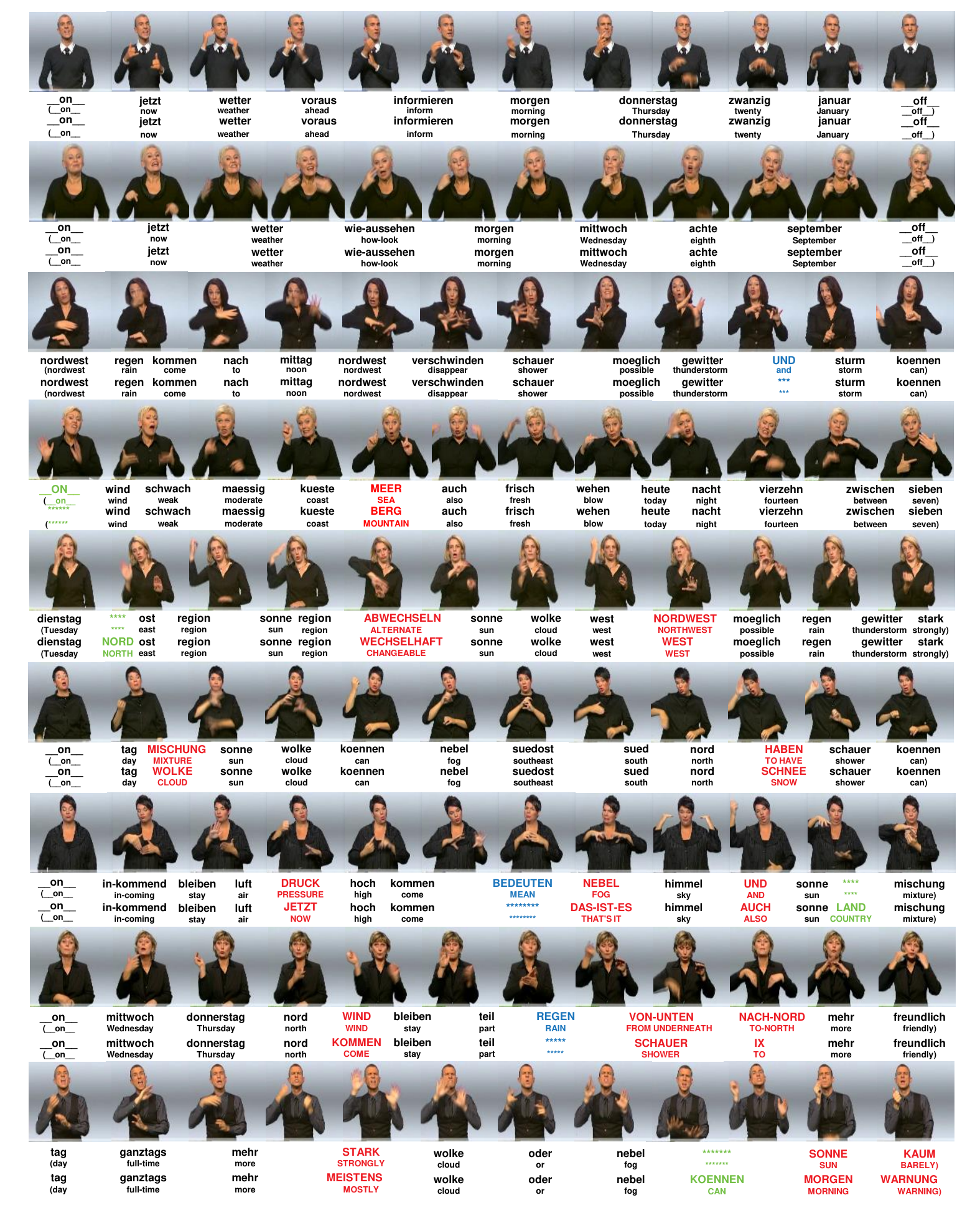}
    \end{center}
    \caption{Recognition results on full videos of the RWTH-PHOENIX-Weather-2014 dataset. \textcolor{green}{Deletion}, \textcolor{blue}{insertion} and \textcolor{red}{substitution} errors are colored in \textcolor{green}{green}, \textcolor{blue}{blue} and \textcolor{red}{red} respectively. Sentences below each sample are ground truth and then results of SF-Net. Translations to English are single word based. \textunderscore\textunderscore on\textunderscore \textunderscore \ and \textunderscore\textunderscore off\textunderscore\textunderscore \ are starting and ending flags while \** represents absence of glosses. Samples are chosen from the validation and testing sets. }
    \label{fig:fig1s}
\end{figure*}

\begin{figure*}[htb]
    \begin{center}
        \includegraphics[width=1\linewidth]{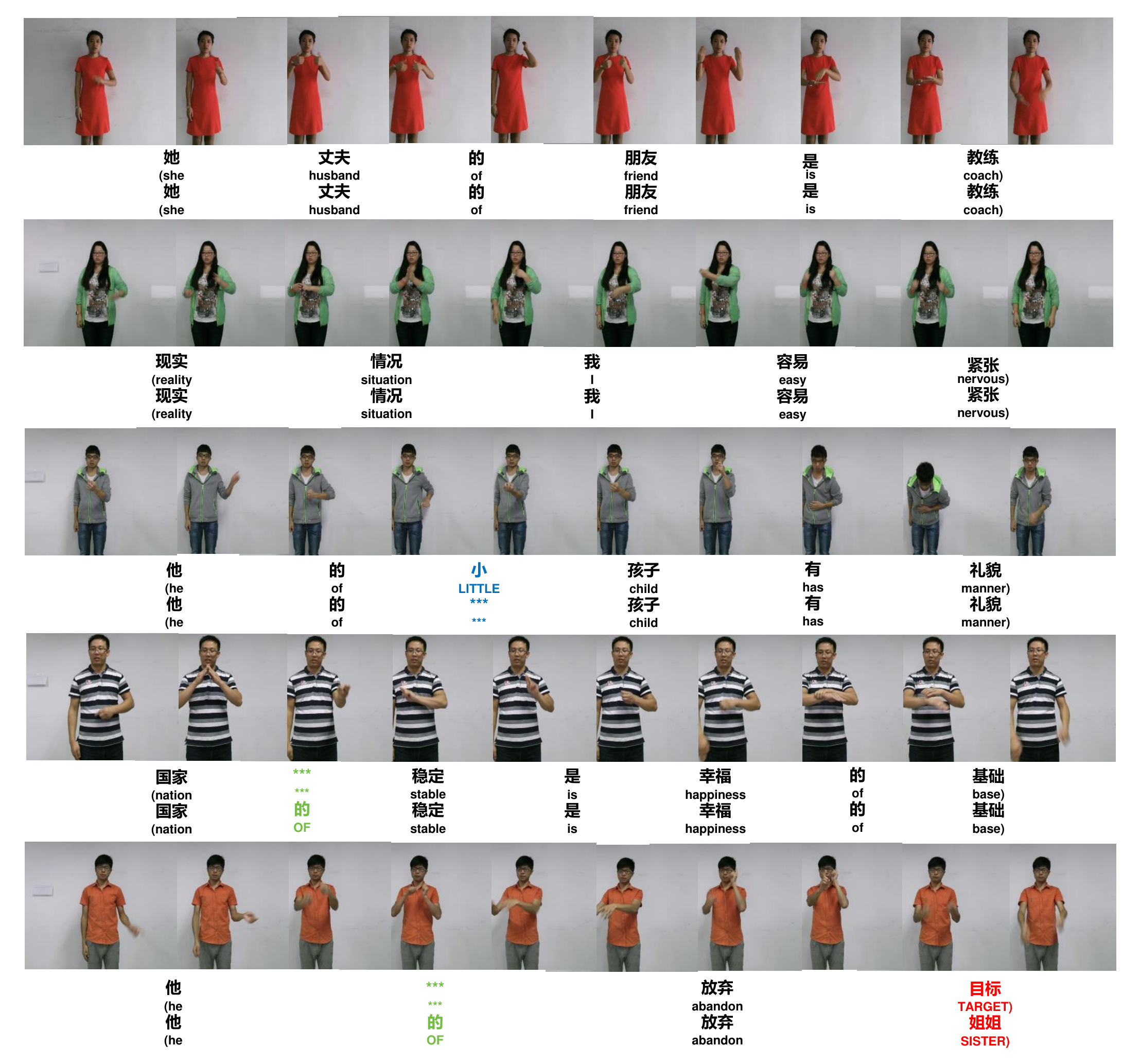}
    \end{center}
    \caption{Recognition results on full videos of the CSL dataset. \textcolor{green}{Deletion}, \textcolor{blue}{insertion} and \textcolor{red}{substitution} errors are colored in \textcolor{green}{green}, \textcolor{blue}{blue} and \textcolor{red}{red} respectively. Sentences below each sample are ground truth and then results of SF-Net. Translations to English are single word based. \** in sentences represents absence of glosses. Samples are chosen from the testing set. }
    \label{fig:fig2s}
\end{figure*}

We show qualitative results of Structured Feature Network (SF-Net) on full videos for the RWTH-PHOENIX-Weather-2014 dataset \cite{rwth} and the Chinese Sign Language (CSL) dataset \cite{lshan} in Figure \ref{fig:fig1s} and Figure \ref{fig:fig2s} respectively. 

The RWTH-PHOENIX-Weather-2014 dataset is richer in expression (vocabulary and sentence length) but less diverse in performance (number of signers and signers' dressings). Sentences in the dataset are unique, so all sentences in the validation and testing set have not been seen by the network during training. We can see that, although the training set is relatively small (compared to other sequence to sequence tasks), the proposed SF-Net is able to recognize running glosses in very long sequence (more than 200 frames). Also, although false recognitions exist, they do not show to have affected other recognition in the sentence. This has demonstrated the robustness of SF-Net. Moreover, many of the false recognition made by SF-Net are close to the ground truth (e.g., northwest (northwest) to west (west), abswchseln (alternate) to wechselhaft (changeable)). They have little impact on understanding the whole sentence. However, bad cases (last 3 samples) also exist. Many of these cases are caused by infrequent glosses (e.g., kaum (barely) which only appears 41 times and druck (pressure) which only appears 95 times in all 65,227 glosses in the training set) and out-of-vocabulary glosses (e.g., noch-nord (to-north) and von-unten (from-underneath)). Note that when there are out-of-vocabulary glosses, adjacent recognitions may be affected. This is because unseen signing patterns can introduce uncertainty in alignment inference.

The CSL dataset contains less sentences but is much more diverse in performance, as it contains more signers and has not unified their dressings. We can see that, the proposed SR-Net is very capable in recognizing seen glosses even when they are signed by unseen signers who dressed different clothes (we have ensured signers in the training and testing sets do not have overlaps). Most of false recognitions are related to prepositions (e.g. gloss `of') that have no influence on the meaning of the whole sentence and can be optionally removed in practice. We did not find many bad cases considering the low word error rate (WER) of 3.8\%. Some (the last sample) may be caused by irregular signing of glosses performed by the signer.

\end{appendix}

\end{document}